# Automatic Coral Detection with YOLO: A Deep Learning Approach for Efficient and Accurate Coral Reef Monitoring

Younes OUASSINE[1], Jihad ZAHIR[1], Noël CONRUYT[2], Mohsen KAYAL[3], Philippe A. MARTIN[2], Eric CHENIN[4], Lionel BIGOT[2], Regine VIGNES LEBBE[5]

[1] Computer Science Department, LISI, Cadi Ayyad University, Morocco
younes.ouaasine99@gmail.com, j.zahir@uca.ac.ma,

[2] EA2525 LIM, I.T. Department, University of La Réunion, 97400 Saint-Denis, France
{noel.conruyt,philippe.martin,Lionel.Bigot}@univ-reunion.fr

[3] ENTROPIE, IRD, IFREMER, CNRS, University of La Reunion, University of New Caledonia, Noumea, New Caledonia
mohsen.kayal@ird.fr

[4] Institut de Recherche pour le Développement (IRD), France
eric.chenin@ird.fr

[5] Institut de Systématique, Evolution, Biodiversité (ISYEB), SU, MNHN, CNRS, EPHE, UA - CP 48, 57 rue Cuvier, 75005 Paris, France
regine.vignes_lebbe@sorbonne-universite.fr

**Abstract.** Coral reefs are vital ecosystems that are under increasing threat due to local human impacts and climate change. Efficient and accurate monitoring of coral reefs is crucial for their conservation and management. In this paper, we present an automatic coral detection system utilizing the You Only Look Once (YOLO) deep learning model, which is specifically tailored for underwater imagery analysis. To train and evaluate our system, we employ a dataset consisting of 400 original underwater images. We increased the number of annotated images to 580 through image manipulation using data augmentation techniques, which can improve the model's performance by providing more diverse examples for training. The dataset is carefully collected from underwater videos that capture various coral reef environments, species, and lighting conditions. Our system leverages the YOLOv5 algorithm's real-time object detection capabilities, enabling efficient and accurate coral detection. We used YOLOv5 to extract discriminating features from the annotated dataset, enabling the system to generalize, including previously unseen underwater images. The successful implementation of the automatic coral detection system with YOLOv5 on our original image dataset highlights the potential of advanced computer vision techniques for coral reef research and conservation. Further research will focus on refining the algorithm to handle challenging underwater image conditions, and expanding the dataset to incorporate a wider range of coral species and spatio-temporal variations.

**Keywords:** Machine Learning, Deep Learning, Underwater ecosystems, Corals, Object Detection, YOLO.

# 1 INTRODUCTION

Coral reefs are among the most biodiverse and productive ecosystems on Earth, supporting an extraordinary array of marine life and providing critical resources to millions of people worldwide [1]. These magnificent underwater structures play a vital role in maintaining the balance of marine biodiversity and serve as natural barriers against coastal erosion and storm surge [2,3]. However, coral reefs face unprecedented threats, primarily due to local anthropogenic impacts and climate change, and are declining, with consequences for biodiversity and coastal societies [4,5,12,13].

The degradation and loss of coral reefs not only endanger marine species but also jeopardize the livelihoods of coastal communities dependent on these vulnerable ecosystems. Effective conservation and management of coral reefs require comprehensive monitoring efforts to assess ecosystem health, identify major drivers of species dynamics, and respond swiftly to changes or disturbances [14,15,16]. Traditional manual survey methods, though essential, are time-consuming, labor-intensive, and often limited in scale, making them insufficient to address the challenges faced by coral reef ecosystems in the present era of accelerating environmental change [6]. Recent advancements in computer vision and deep learning have paved the way for novel approaches to automate coral reef monitoring [7]. The integration of these technologies allows for the development of robust and efficient systems capable of automatic coral detection in large-scale underwater image and video datasets. In this paper, we propose a state-of-the-art solution for automatic coral detection, employing the popular You Only Look Once (YOLO) object detection algorithm [11].

The main aim of this research is to show how YOLOv5, known for its real-time object detection capabilities, can be used to effectively and accurately detect coral colonies in underwater imagery. By using YOLOv5 instead of other deep learning techniques, our approach aims to surpass traditional methods and enhance the efficiency, scalability, and accuracy of coral reef monitoring. In the Results and Analysis section, we present a complete analysis of our automatic coral detection system, detailing the key components, architecture, and data preparation. We also discuss the challenges associated with coral detection in underwater environments, such as lighting variations and complex background structures. To address these challenges, we describe the pre-processing techniques applied to the input imagery to improve the overall performance of the YOLOv5 model. Overall, this paper aims to contribute to the growing body of research in the field of imagery tools and artificial intelligence algorithms applied to coral reef monitoring and conservation [6,7].

# 2 RELATED WORKS

The authors of [8] Focuses on the classification of bleached and unbleached corals using visual vocabulary, which combines spatial, texture, and color features. The proposed methodology includes feature extraction techniques and classifiers, with a bag of features (BoF) and a linear kernel of Support Vector Machine (SVM) achieving the highest accuracy of 99.08% for binary classification and 98.11% for multi-class classification.

The authors in [9] Compares two supervised machine learning methods for detecting and recognizing coral reef fish in underwater videos. The authors present the Deep Learning method and the HOG+SVM method and evaluate their performance. The paper discusses the use of histograms of oriented gradients (HOG) for feature extraction and support vector machines (SVM) for classification in the HOG+SVM method. The Deep Learning method utilizes a deep neural network for both feature extraction and classification. The authors compare the F-measure of both methods on a dataset of underwater videos, for the HOG+SVM approach, the F-measure ranges from 0.28 to 0.49, while for the Deep Learning method, it ranges from 0.62 to 0.65.

The authors of [10] Introduces a new method for detecting and assessing the health of coral reefs using underwater images and videos. The researchers found that previous studies focused mainly on coral image classification and lacked coral health detection. To address this gap, they developed a hardware-based autonomous monitoring system for coral-reef health detection. The proposed method, called MAFFN-YOLOv5, consists of a backbone, neck, and head network architecture. The authors obtained 0.83 in the precision measure for a dataset of 3049 high-quality images.

## 3 DATASETS

We used a dataset of 580 underwater images specifically collected for coral detection. The dataset includes 400 original images, which were augmented through image treatment to increase training data diversity. The images were captured using underwater cameras during research expeditions to various coral reef locations in the Reunion and Scattered Islands. These expeditions aimed to cover a wide range of coral reef health and habitats, depths, and lighting conditions, ensuring a comprehensive representation of underwater environments. The dataset includes different coral species, colony sizes, and orientations, providing a realistic depiction of the coral reef ecosystem.

Public link to the dataset: https://drive.google.com/file/d/1YsqGLyAZ4QRZkUJHs8VsRrfj6bK7CG1C/view

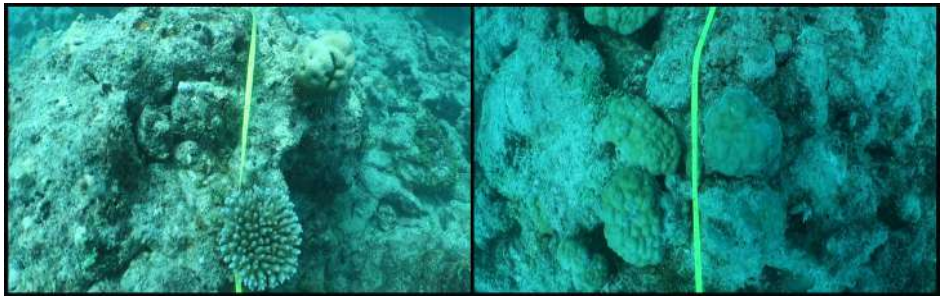

**Fig. 1.** Sample images from the dataset

For training and evaluating our automatic coral detection system, we manually annotated the images using the Label Studio tool. The annotation process involved drawing bounding boxes around individual coral colonies and assigning them the label "coral". This annotation scheme simplifies the task by focusing on identifying coral colonies specifically.

44

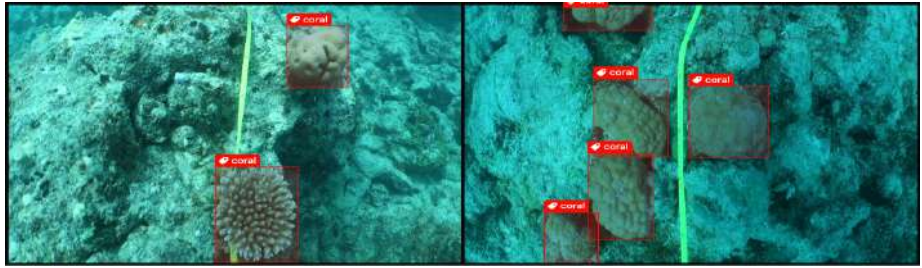

**Fig. 2.** Sample annotated images from the dataset

## 4   YOLOv5

YOLOv5, like other neural networks designed for image processing, is built upon convolutional layers. The architecture of YOLOv5 can be divided into three distinct parts, as illustrated in Figure 3. These parts can be considered as sets of layers that form functional units. Each part serves a specific role and is composed of modified versions of existing network components found in the literature. These modified layers were carefully selected and tailored to suit the requirements of YOLOv5's architecture.

- Backbone: The initial component of the network is referred to as the backbone. Its primary function is to optimize the network by enhancing the gradient descent process. This crucial segment enables the network to achieve fast inference times, facilitating real-time applications.
- Neck: The neck component in the model generates feature pyramids, which aid in generalizing objects across various scales. These feature pyramids enhance the model's ability to identify objects of different sizes and scales, improving its robustness. YOLOv5 incorporates PANet, a feature pyramid approach, which contributes to superior predictions, increased accuracy, and improved performance.
- Head: The head model plays a crucial role in the final stage of detection. It is responsible for generating the output vectors that contain class probabilities, objectivity scores, and bounding boxes by utilizing anchor boxes. These anchor boxes serve as reference templates that aid in accurately localizing and classifying objects within the input data.

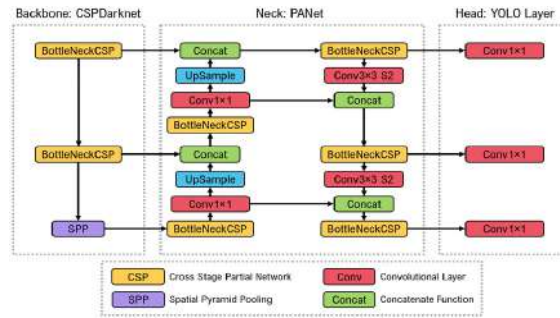

**Fig. 3.** YOLOv5 architecture

# 5 RESULTS AND ANALYSIS

YOLOv5 is available in different variants, designated by the letters N, S, M, and L, representing NANO, Small, Medium, and Large, respectively. These variants differ in the size of the model architecture, which impacts performance and detection speed. In general, the larger variants (M and L) tend to offer higher accuracy but slower inference times, while the smaller variants (N and S) sacrifice some accuracy for faster performance. The table presents a comparison of various models in the YOLOv5 family specifically designed for coral detection tasks. Each model is evaluated based on its size, precision, recall, mean average precision, and inference time. Here is a description of each column:

**Table 1 :** Comparison of YOLOv5 models for coral detection

| Model | Size | Precision | Recall | m Average Precision | Inference time |
|---|---|---|---|---|---|
| **Yolov5n** | 3.9MB | 0.534 | 0.46 | 0.461 | 50.2ms |
| **Yolov5s** | 14.4MB | 0.558 | 0.445 | 0.452 | 83.7ms |
| **Yolov5m** | 42.2MB | 0.598 | 0.481 | 0.47 | 178.5ms |
| **Yolov5l** | 92.8MB | 0.599 | 0.486 | 0.474 | 319.5ms |

- Model: This column displays the different YOLOv5 models specifically tailored for coral detection that are being compared.
- Size: This column indicates the file size of each YOLOv5 model, which represents the amount of storage required for the model.
- Precision: Precision measures the accuracy of the model's predictions for coral detection. It calculates the ratio of true positive predictions to all positive predictions made by the model.
- Recall: Recall quantifies the ability of the model to correctly identify coral instances in the dataset. It represents the ratio of true positive predictions to all actual positive instances present.
- mean Average Precision: mean Average Precision is a widely used metric in object detection tasks, including coral detection. It calculates the average precision across different levels of detection confidence thresholds. Higher values indicate better overall performance in terms of precision and recall.
- Inference Time: This column represents the time taken by each YOLOv5 model to process an input image and generate the corresponding coral detection results. It is measured in milliseconds (ms).

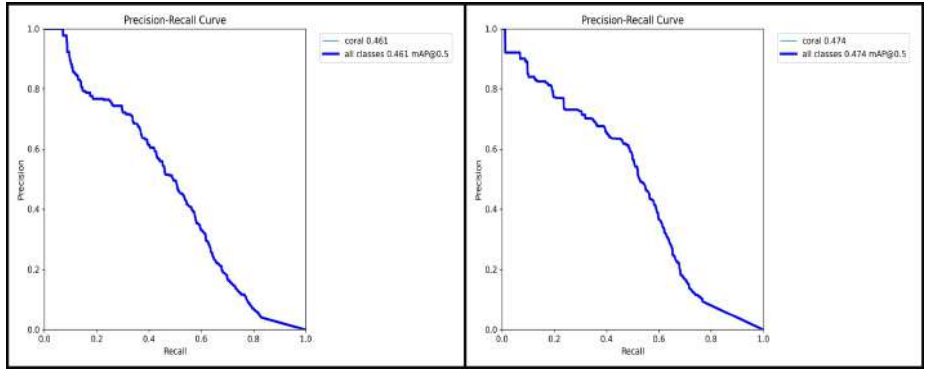

**Fig. 4.** precision and recall curves for yolov5n and yolov5l models respectively from left to right

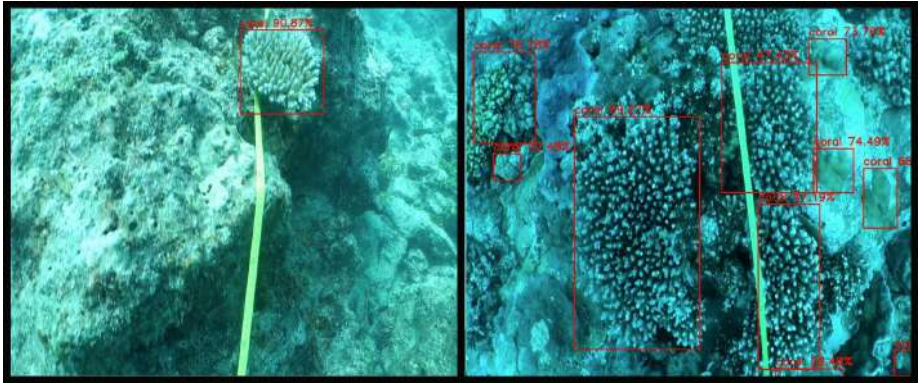

**Fig. 5.** Examples of coral detection results

## 6      CONCLUSION

In this paper, we have presented an automatic coral detection system using the You Only Look Once (YOLO) algorithm, a state-of-the-art deep learning approach for efficient and accurate object detection. Our proposed system addresses the need for scalable and accurate coral reef monitoring, a critical task for understanding and conserving these vulnerable ecosystems. Our approach is specifically tailored to handling difficult image data, which often contains corals with various shapes and sizes, as well as images featuring an uneven distribution of corals.

While the utilization of our dataset comprising 580 manually annotated underwater images with precise bounding boxes has been beneficial for training and evaluating the automatic coral detection system, it is important to acknowledge some limitations. Currently, our approach primarily focuses on images with a low concentration of corals. To improve the system's effectiveness in detecting corals in images with a high concentration of corals, future work involves annotating more images and data specifically targeting such scenarios. This expansion of annotated data will enhance the model's ability to accurately detect and analyze densely populated coral areas. By automating coral detection with YOLOv5, we provide a means to gather timely and accurate data, enabling a deeper understanding of coral reef dynamics and facilitating prompt actions in response to threats. This enhanced understanding allows us to develop targeted conservation strategies, allocate resources efficiently, and implement adaptive management practices. Our research contributes to the broader field of computer vision and deep learning in marine biology and environmental sciences, fostering innovation and collaboration to tackle pressing sustainability challenges. By successfully applying YOLOv5 to coral detection, we pave the way for scalable and cost-effective monitoring solutions, thereby supporting long-term coral reef resilience and biodiversity preservation.